\documentclass{article}
\usepackage{spconf,amsmath,graphicx,amsmath,amsthm,paralist,amssymb,amsfonts}
\usepackage{tikz}
\usepackage{graphicx,comment,xfrac}

\title{A GMM based algorithm to generate point-cloud and its application to neuroimaging}
\name{Liu Yang and Rudrasis Chakraborty}
\address{University of California, Berkeley, USA}
\begin{document}
%
\maketitle

\begin{abstract}
Recent years have witnessed the emergence of 3D medical imaging techniques with the development of 3D sensors and technology. Due to the presence of noise in image acquisition, registration researchers focused on an alternative way to represent medical images. An alternative way to analyze medical imaging is by understanding the 3D shapes represented in terms of point-cloud. Though in the medical imaging community, 3D point-cloud processing is not a ``go-to'' choice, it is a ``natural'' way to capture 3D shapes. However, as the number of samples for medical images are small, researchers have used pre-trained models to fine-tune on medical images. Furthermore, due to different modality in medical images, standard generative models can not be used to generate new samples of medical images. In this work, we use the advantage of point-cloud representation of 3D structures of medical images and propose a Gaussian mixture model-based generation scheme. Our proposed method is robust to outliers. Experimental validation has been performed to show that the proposed scheme can generate new 3D structures using interpolation techniques, i.e., given two 3D structures represented as point-clouds, we can generate point-clouds in between. We have also generated new point-clouds for subjects with and without dementia and show that the generated samples are indeed closely matched to the respective training samples from the same class. 
\end{abstract}
\section{Introduction} \label{intro}

Since the inception of medical image analysis, researchers have been using 3D imaging to capture structure of the brain. Throughout the last decade, this community has seen the emergence of deep learning due to its power to capture the local structure. One of the major hurdle in medical imaging is the lack of samples. In the era of deep learning, one needs to have a lot of training samples in order to learn a deep network. This is the main reason researchers failed short to use deep learning models for medical imaging data and resort to alternative approaches like transfer learning \cite{shin2016deep}. 

This motivates researchers to generate more training samples. Generative models like GAN \cite{goodfellow2014generative, arjovsky2017wasserstein}, flow based models \cite{kingma2018glow} have been immensely popular to generate high resolution natural images. But for medical imaging data, due to the different modalities, generating images is challenging. Furthermore, medical images are three dimensional, hence convolution based generative models are computationally expensive. This motivates us to explore alternative of images and in general alternatives of 3D structure representation. Point-cloud is an efficient way to represent 3D structures \cite{qi2017pointnet,qi2017pointnet++} because of its important geometric properties. Due to the lack of a smooth topology, standard convolution can not be applied on point-cloud. One of the popular approaches to do point convolution \cite{zhou2018voxelnet} is to divide the point-cloud into voxels and then extract some features using 3D convolution. However, this method suffers from the possible sparsity of point-clouds which results in multiple empty voxels. One possible solution is to use multi-layer perceptron (MLP) to extract features from each point \cite{qi2017pointnet} or from a local neighborhood around each point \cite{qi2017pointnet++}. Unfortunately, all these methods are susceptible to noise and hence is not robust to outliers which makes them incapable to efficiently deal with measurement and registration errors common in medical images.

In recent years, several researchers \cite{chakraborty2017statistics,muralidharan2012sasaki} have proposed methods to do discrimination between subjects with and without dementia. Some of the popular approaches include using the 3D volume of the region of interest (ROI) and analyzing the shape of the anatomical structures of interest. In order to do that, the researchers either proposed techniques to map the 3D volume in high dimensional space \cite{chakraborty2017statistics} or mapped the shape of the anatomical structure on the complex projective space, i.e., Kendall's shape space \cite{muralidharan2012sasaki}. 

In this work, we propose a Gaussian mixture model (GMM) based point cloud generated scheme, we have shown that our proposed algorithm can be applied to generated 3D structures like corpus callosum, which is one of the most important region affected by neurological disorders, e.g., dementia, tremor etc.. Researchers have been used GMM to model point-cloud before \cite{jian2010robust}, we extend that idea to generate point-clouds and also we propose an interpolation scheme to generate CC shapes on a geodesic between two given CC shapes. We tested our proposed approach on publicly available OASIS dataset \cite{fotenos2005normative}. Experimental results have shown that our proposed methods can act as a simple yet effective point-cloud generation technique. 

The salient features of our proposed method are: \begin{inparaenum}[\bfseries (1)] \item The proposed GMM based 3D shape model is robust to outliers. \item We propose a simple scheme to generate point-clouds. \item Using experiments, we show that our proposed scheme can generate point-clouds conditioned on the class, i.e., can generate point-clouds for subjects with and without dementia. \end{inparaenum}

\section{GMM based point-cloud generation} \label{theory}

In this work, we propose a point-cloud generation algorithm. Our algorithm consists of three key steps \begin{inparaenum} \item fit a Gaussian mixture model (GMM) based on Expectation-Maximization (EM), \item choose the number of components of the GMM based on AIC based criterion, \item draw samples from the learned GMM to generate point-cloud. \end{inparaenum} Below we will describe the key components of our proposed scheme.

{\bf EM step to fit a GMM:} Given a point-cloud $X = \left\{\mathbf{x}_i\right\}_{i=1}^N \subset \mathbf{R}^3$ and the desired number of components, $K$ (we will learn the value of $K$ next), we design and Expectation-Maximization (EM) algorithm as follows. Let the learned GMM be represented by $\left\{w_j, \boldsymbol{\mu}_j, \Sigma_j\right\}_{j=1}^K$, where $\forall j, w_j \geq 0$ and $\sum_{j=1}^K w_j = 1$, $\left\{\boldsymbol{\mu}_j\right\} \subset \mathbf{R}^3$ and  $\forall j, \Sigma_j \in P_3$, where $P_3$ is the space of symmetric positive definite matrices. 

We run $K$-means algorithm on $X$ to initialize $\left\{\boldsymbol{\mu}_j\right\}, \left\{\Sigma_j\right\}$ using the mean and covariance of the clusters. We initialize $w_j = \sfrac{N_j}{N}$, where $N_j$ is the number of points in $j^{th}$ cluster. 

{\it E step:} Compute the probability of $\mathbf{x}_i$ belong to $j^{th}$ cluster (denoted by $\gamma_{ij}$) as:
\begin{align}
\gamma_{ij} = \frac{w_j f\left(\mathbf{x}_i|\boldsymbol{\mu}_j, \Sigma_j\right)}{\sum_{j=1}^K w_j f\left(\mathbf{x}_i|\boldsymbol{\mu}_j, \Sigma_j\right)},
\end{align}  
where, 
\begin{align*}
f\left(\mathbf{x}_i|\boldsymbol{\mu}_j, \Sigma_j\right) = \frac{\exp\left(-\frac{1}{2}\left(\mathbf{x}_i - \boldsymbol{\mu}_j\right)^t\Sigma_j^{-1}\left(\mathbf{x}_i - \boldsymbol{\mu}_j\right)\right)}{\left(2\pi \text{det}\left(\Sigma_j\right)\right)^{3/2}}.
\end{align*}

{\it M step:} Update the parameters $\left\{w_j, \boldsymbol{\mu}_j, \Sigma_j\right\}_{j=1}^K$ as:
\begin{align*}
w_j = \frac{\sum_{i=1}^{N_j}\gamma_{ij}}{\sum_{j=1}^K \sum_{i=1}^{N_j} \gamma_{ij}}
\end{align*}

\begin{align*}
\boldsymbol{\mu}_j = \sum_{i=1}^{N_j} \frac{\gamma_{ij}}{\sum_{k=1}^{N_j} \gamma_{kj}} \mathbf{x}_i
\end{align*}

\begin{align*}
\Sigma_j = \sum_{i=1}^{N_j} \frac{\gamma_{ij}}{\sum_{k=1}^{N_j} \gamma_{kj}} \left(\mathbf{x}_i - \boldsymbol{\mu}_j\right)\left(\mathbf{x}_i - \boldsymbol{\mu}_j\right)^t
\end{align*}

We repeat $E$ and $M$ step until convergence. After this algorithm we output the GMM $\left\{w_j, \boldsymbol{\mu}_j, \Sigma_j\right\}_{j=1}^K$.

{\bf Choose the number of components $K$:} Let us denote the GMM with $k$ components resulted from the previous step as $\mathcal{N}^k$. Let us denote the feasibility set for $k$ to be $\mathcal{I}$. We can compute the AIC score for each $\mathcal{N}^k$ where $k \in \mathcal{I}$. Let the AIC scores be denoted by $\left\{\text{AIC}_k\right\}_{k\in \mathcal{I}}$. Then, we normalize the AIC scores to get $\exp\left(\left(\text{AIC}_m - \text{AIC}_k\right)/2\right)$, where $\text{AIC}_m$ is the minimum AIC scores. Let the normalized scores be denoted by $\left\{nA_k\right\}$. We use a threshold of $0.01$ on these scores and return the final GMM model as
\begin{align}
\mathcal{N}^* := \sum_{k \in \mathcal{I}, nA > 0.01} p_k \mathcal{N}^k, 
\end{align}
where, $p_k = \frac{nA_k}{\sum_{j \in \mathcal{I}, nA_j > 0.01}}$.

{\bf Draw samples from $\mathcal{N}^*$:} We draw a sample from $\mathcal{N}*$ parametrized by $\left\{p_k, \left\{w^k_j\right\}, \left\{\boldsymbol{\mu}^k_j\right\}, \left\{\Sigma^k_j\right\}\right\}$ as follows. 
\begin{enumerate}
\item Draw a sample from categorical distribution with parameter $\left\{p_k\right\}$. Let the sample be $k$.
\item Draw a sample from categorical distribution with parameter $\left\{w^k_j\right\}$. Let the component be $j$.
\item Draw a sample sample from Gaussian distribution with mean $\boldsymbol{\mu}^k_j$ and covariance matrix $\Sigma_j^k$.
\item Repeat steps (1)-(3) for $N$ times to generate a point-cloud with $N$ points.
\end{enumerate}

This concludes our algorithm to generate point-cloud. Now, we will describe the algorithm to do interpolate between two given point-clouds, $X$ and $Y$ using the point-cloud generation algorithm discussed above. 

{\bf Interpolate between two point-clouds:} Given two point-clouds $X$ and $Y$, denoted by GMMs $\mathcal{N}^X$ and $\mathcal{N}^Y$, with number of components to be $K_1$ and $K_2$ respectively, we interpolate to get a point-cloud $Z$ as follows:
\begin{enumerate}
\item Let $K = \min\left\{K_1, K_2\right\}$. Project $\mathcal{N}_X$ and $\mathcal{N}^Y$ to the nearest $K$ component GMM.
\item Let the parameters be $\left\{w^x_j, \boldsymbol{\mu}^x_j, \Sigma^x_j\right\}$ and  $\left\{w^y_j, \boldsymbol{\mu}^y_j, \Sigma^y_j\right\}$.
\item We identify $\mathbf{w}^x$ and $\mathbf{w}^y$ as points on $\mathbf{S}^{K-1}$ by using square root parametrization \cite{srivastava2007riemannian}. Thus we identify each GMM, i.e., $\mathcal{N}^X$ and $\mathcal{N}^Y$ as a point on the product space $\mathcal{P} :=\mathbf{S}^{K-1} \times \mathbf{R}^{3\times K} \times \left(P_3\right)^K$. 
\item We do interpolation on the product space $\mathcal{P}$ (using the geodesic expression given below) and use the generation algorithm to generate the corresponding interpolated point-cloud. 
\end{enumerate}

{\bf Expression for geodesic on $\mathcal{P}$:} We use the arc-length, $\ell_2$ and \textsf{GL}-invariant distances on $\mathbf{S}^{K-1}$, $\mathbf{R}^{3\times K}$ and $P_3$ respectively. The analytic expression of the (shortest) geodesic is given by:
\begin{align*}
\Gamma_{(\mathbf{w}_1, M_1, \Sigma_1)}^{(\mathbf{w}_2, M_2, \Sigma_2)}(t) =& \left(\frac{\sin(\theta)}{\mathbf{w}_1\sin((1-t)\theta)+\mathbf{w}_2\sin(t\theta)}\right., \\ 
& \left.(1-t)M_1 + tM_2, \right. \\
& \left. M_1^{0.5} \left(M_1^{-0.5}M_2M_1^{-0.5}\right)^t M_1^{0.5}\right), 
\end{align*}
where, $\theta = \arccos(\mathbf{w}_1^t\mathbf{w}_2)$.

In the next section, we will give the data description and the experimental details.

\section{Experimental results} \label{results}
This section consists of the data description followed by the details of experimental validation. 
\ \\
{\bf Data description:} In this section, we use OASIS data \cite{fotenos2005normative} to address the
classification of demented vs. non-demented subjects using
our proposed framework. This dataset contains at least two MR brain
scans of $150$ subjects, aged between $60$ to $96$ years old. For each
patient, scans are separated by at least one year. The dataset
contains patients of both sexes. In order to avoid gender effects, we take MR scans of male patients alone from three visits, which
resulted in the dataset containing $69$ MR scans of $11$ subjects with
dementia and $12$ subjects without dementia. This gives $33$ scans for subjects with dementia and $36$ scans for subjects without dementia. We first compute an atlas
(using the method in \cite{avants2009advanced}) from the $36 (=12
\times 3)$ MR scans of patients without dementia.

\begin{figure*}[!ht]
 \centering
\includegraphics[scale=0.22]{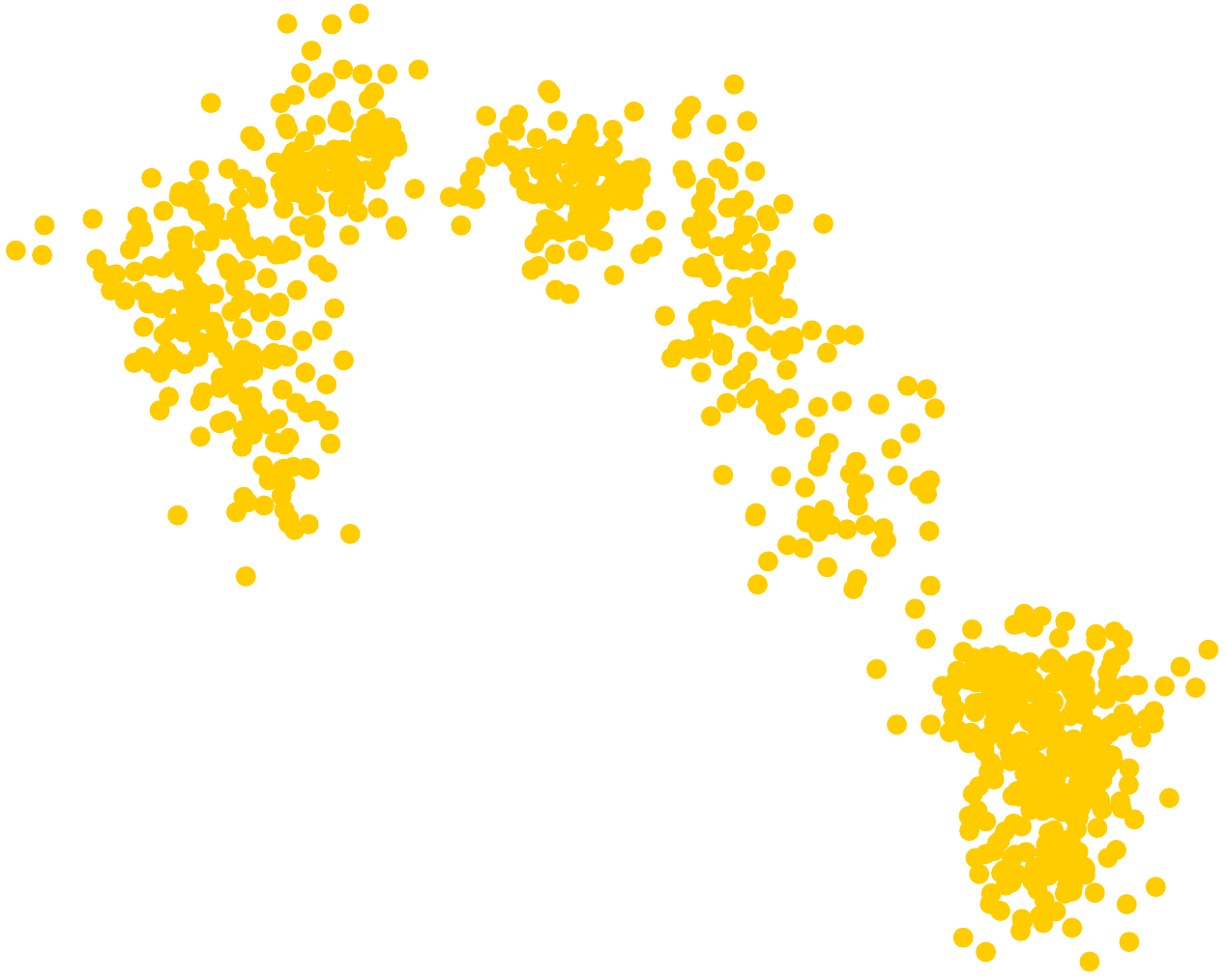}
\includegraphics[scale=0.22]{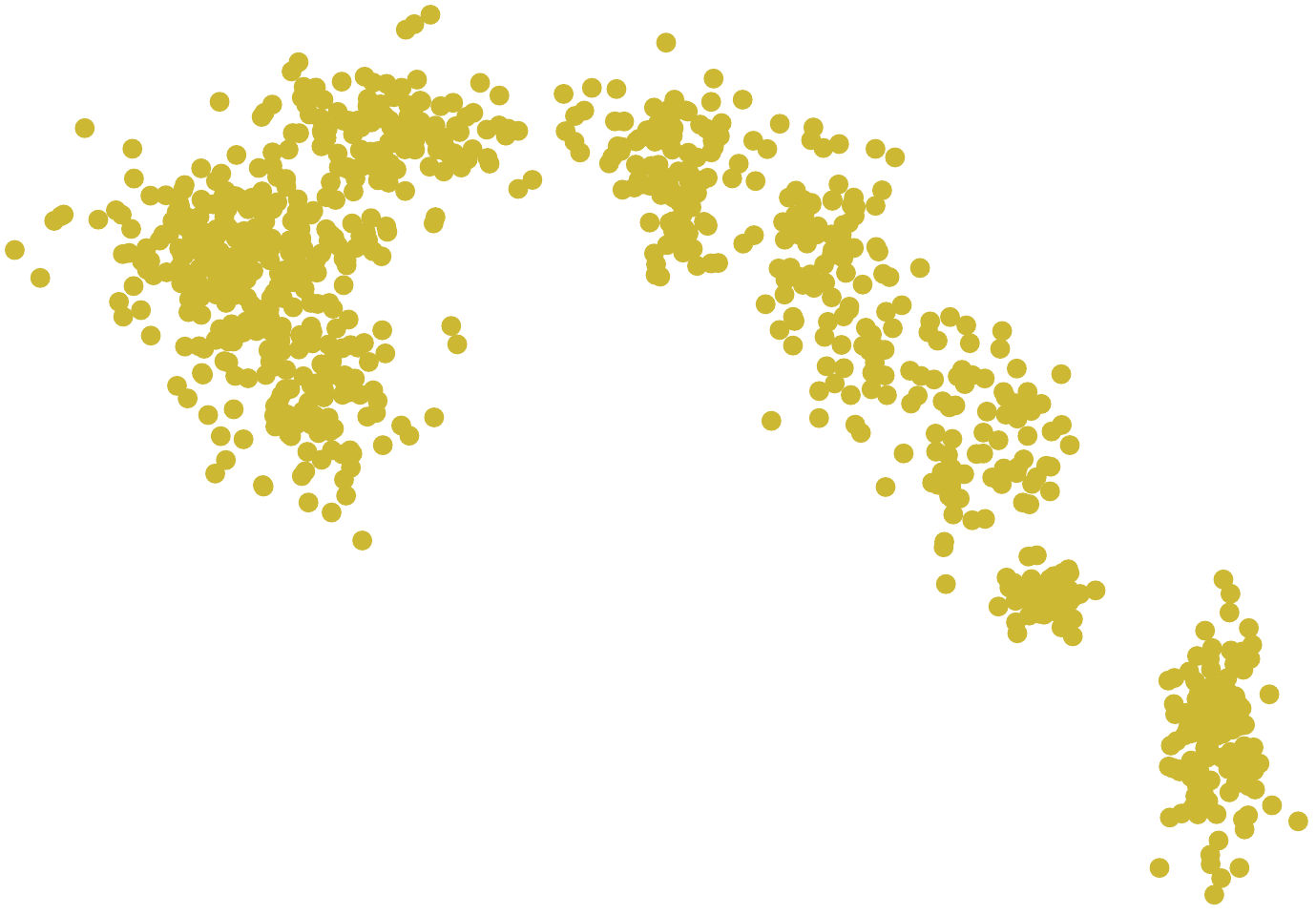}
\includegraphics[scale=0.22]{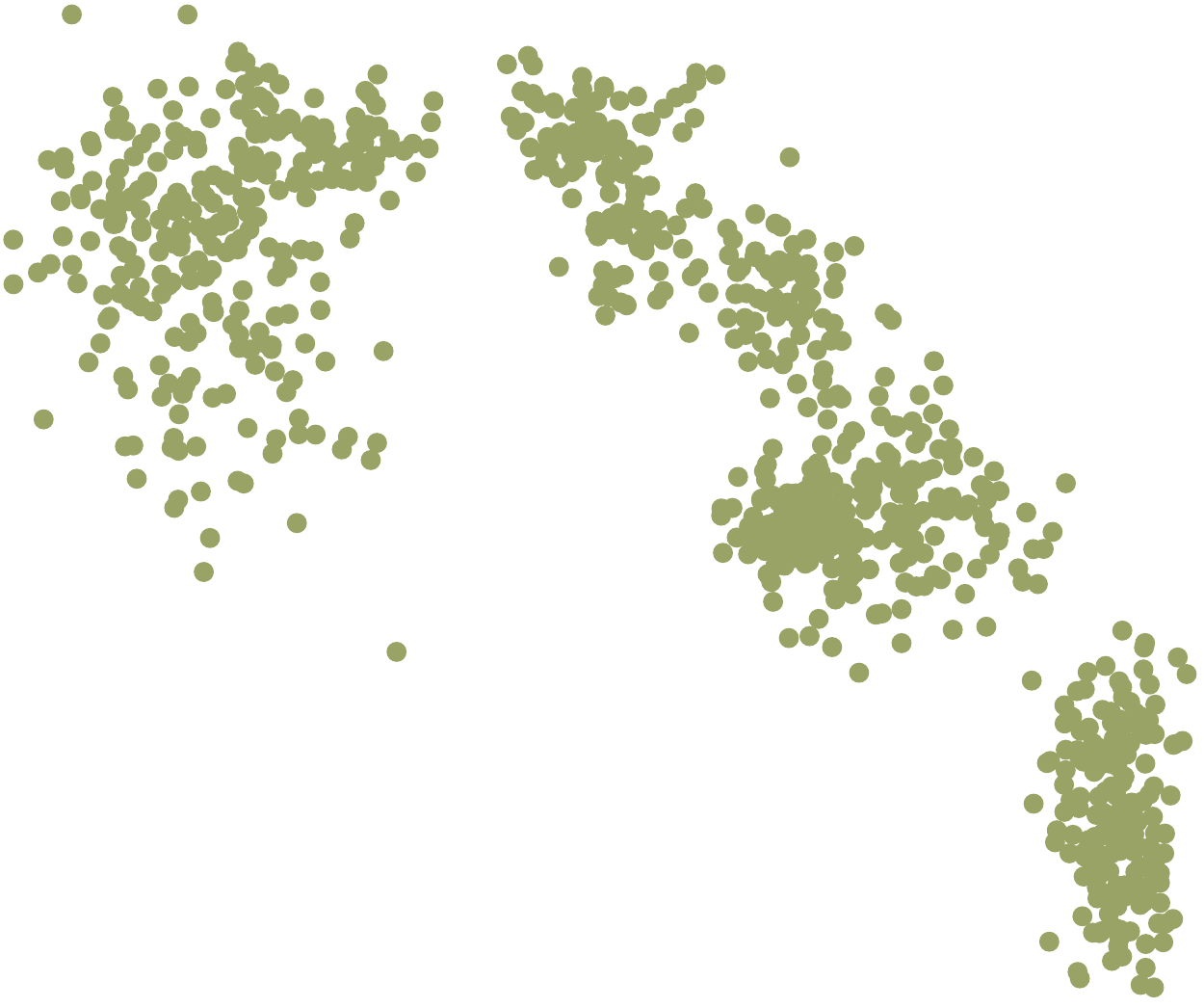}
\includegraphics[scale=0.22]{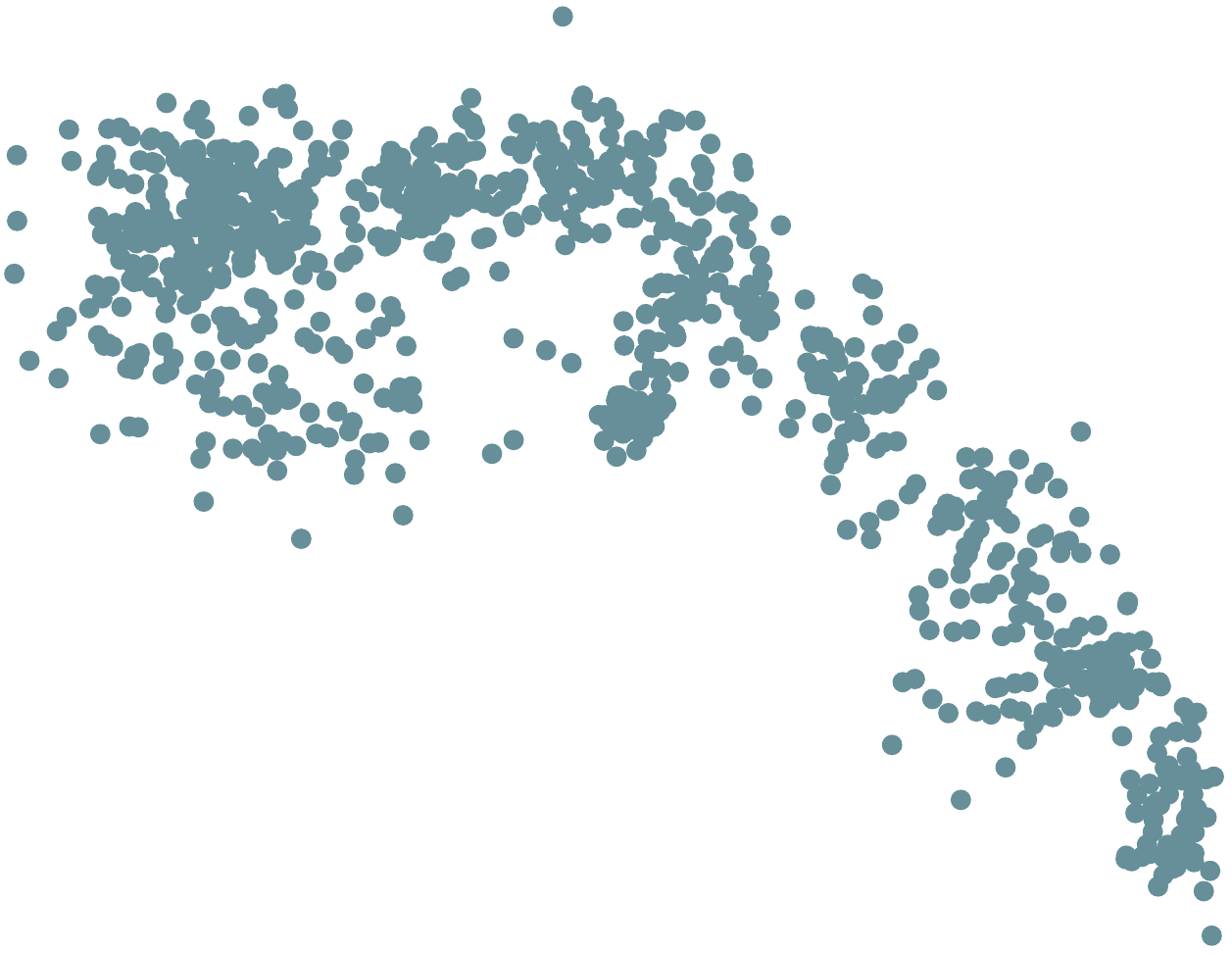}
\includegraphics[scale=0.22]{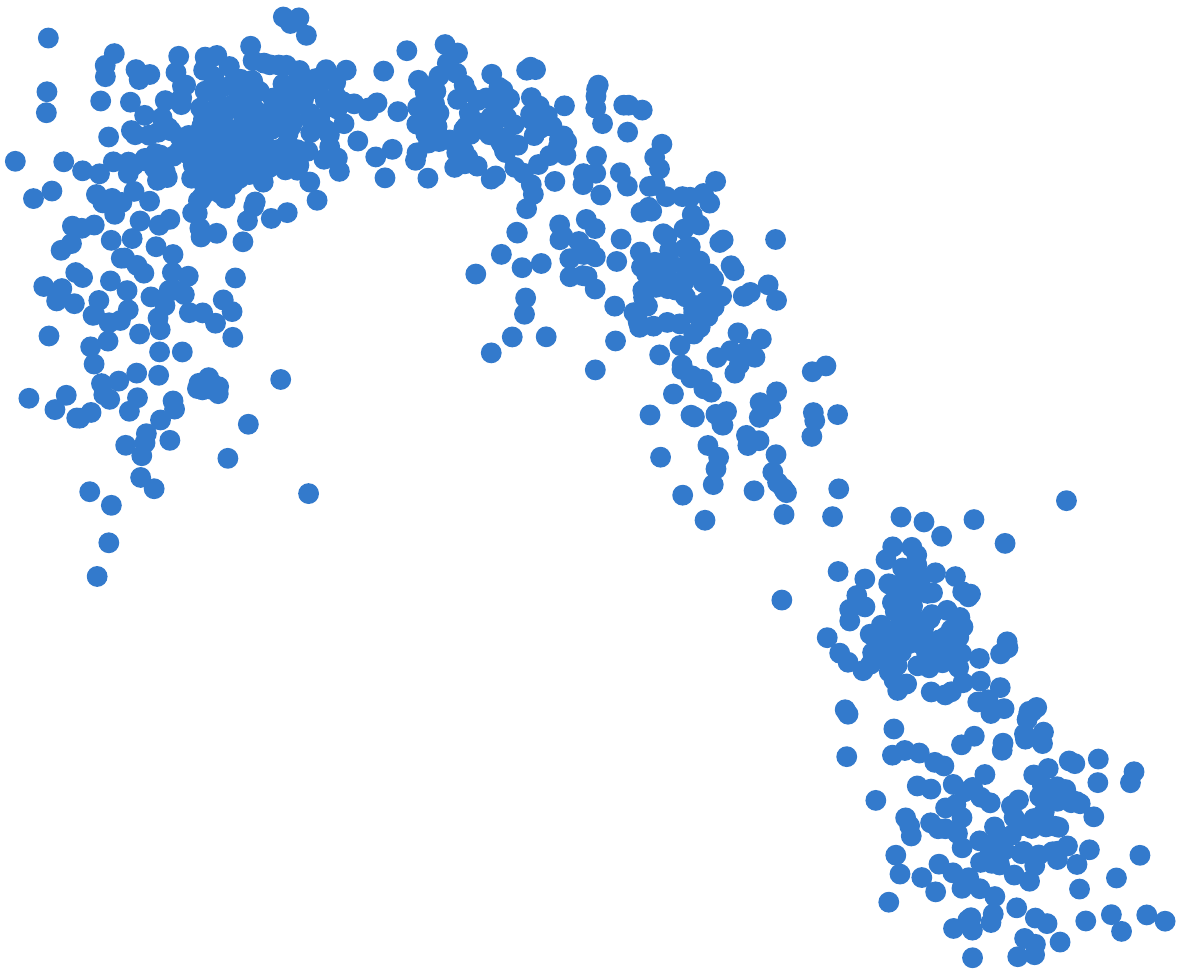}
\includegraphics[scale=0.22]{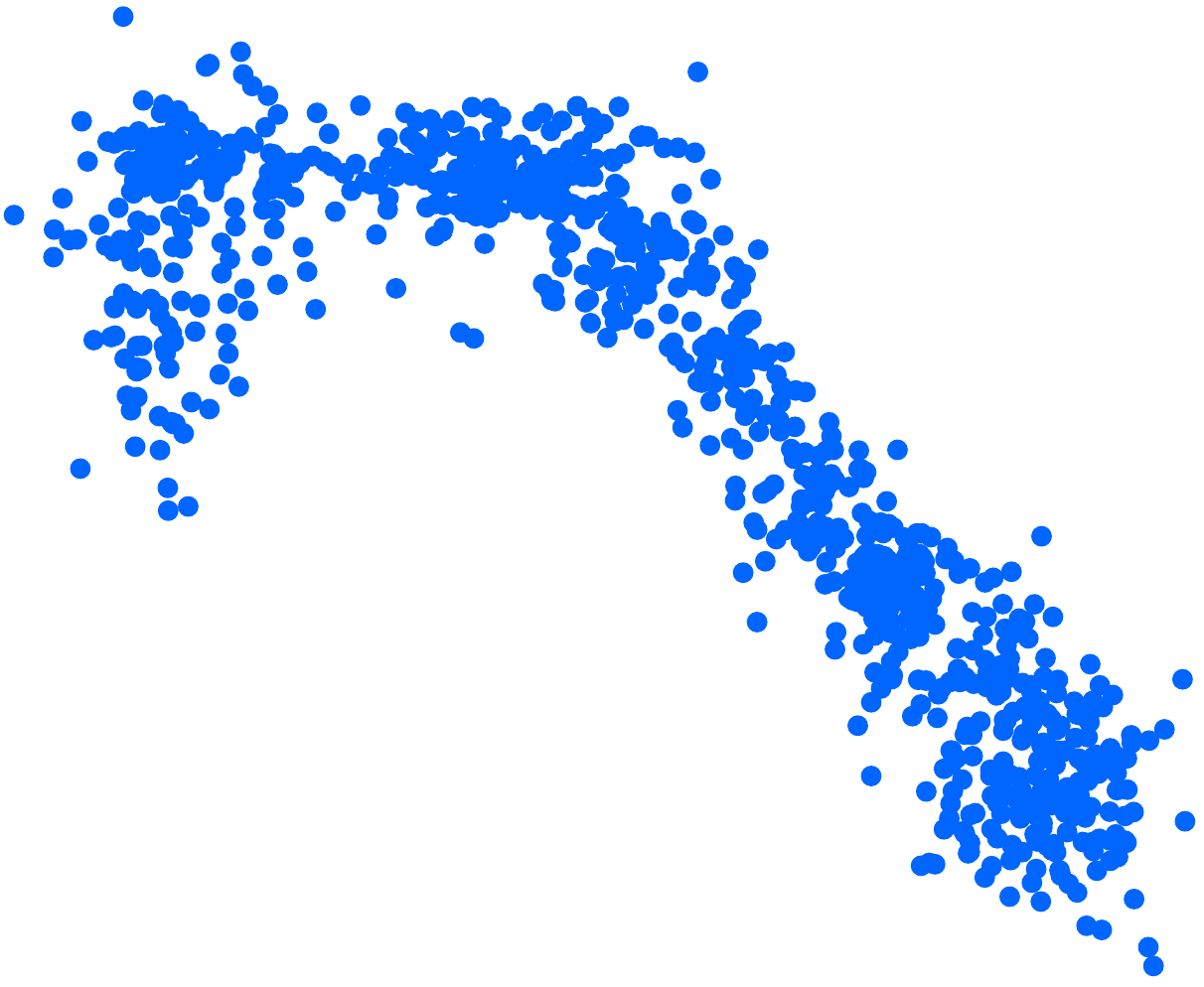}
\caption{Interpolation results for point cloud extracted from CC shapes, the figures are shown with $t=0, 0.2, 0.4, 0.6, 0.8, 1.0$. }
      \label{fig1}
\end{figure*}

\begin{figure*}[!ht]
 \centering
\includegraphics[scale=0.22]{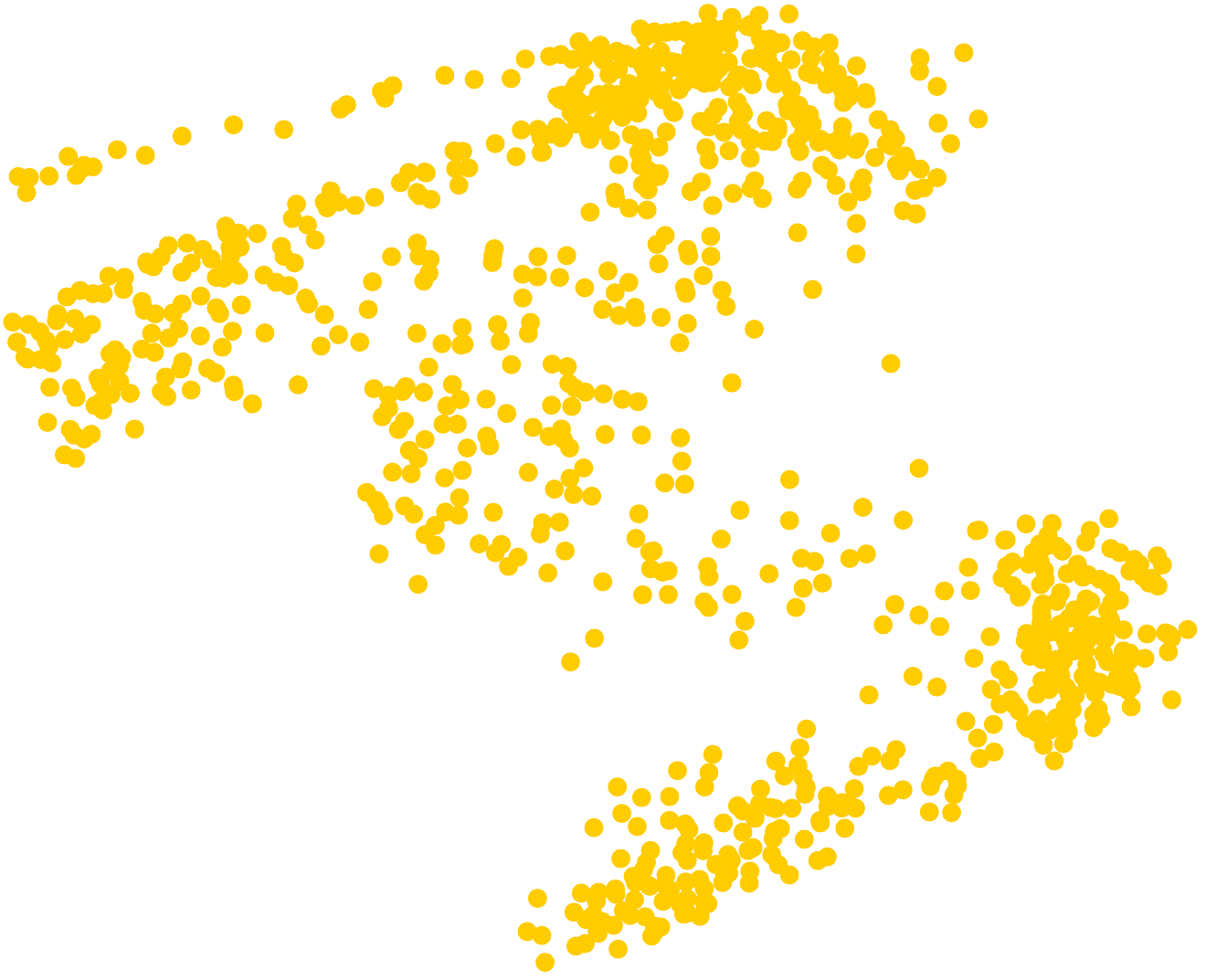}
\includegraphics[scale=0.22]{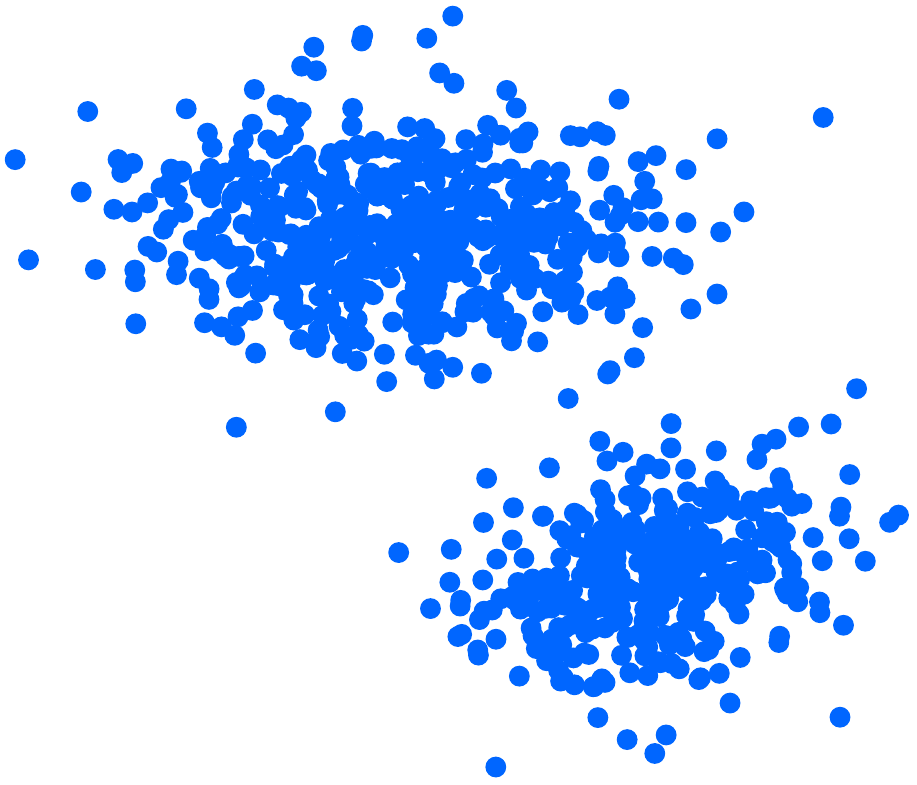}
\includegraphics[scale=0.22]{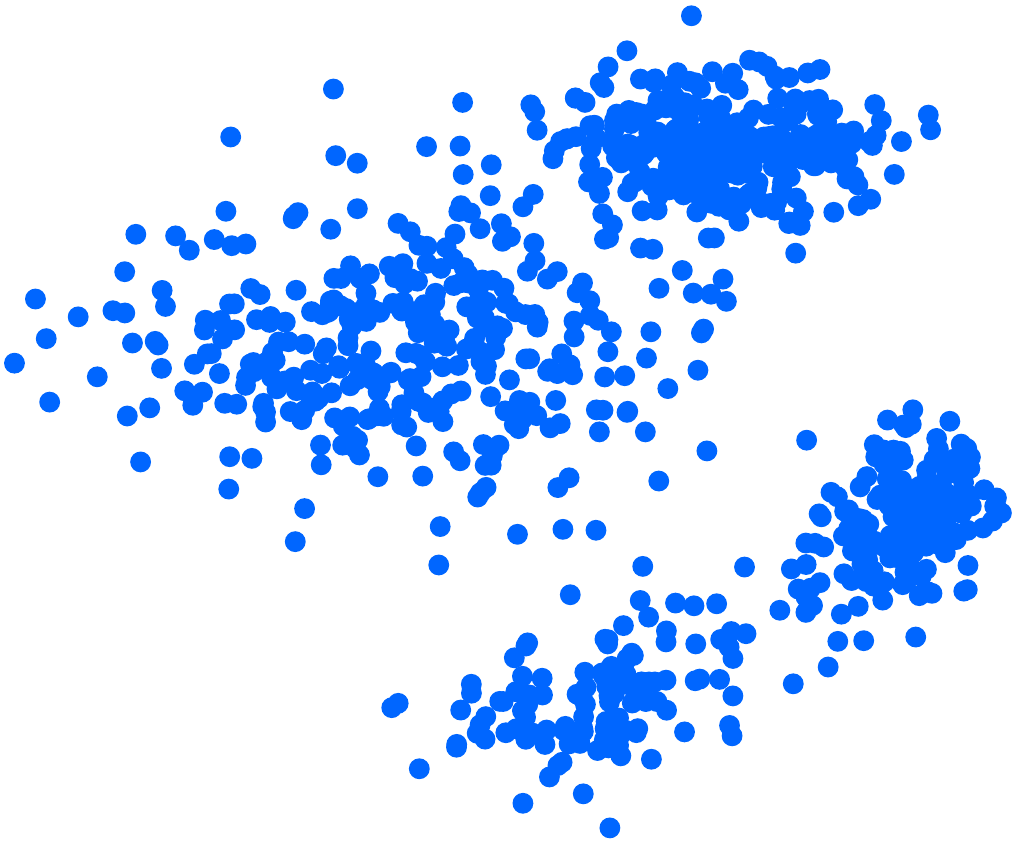}
\includegraphics[scale=0.22]{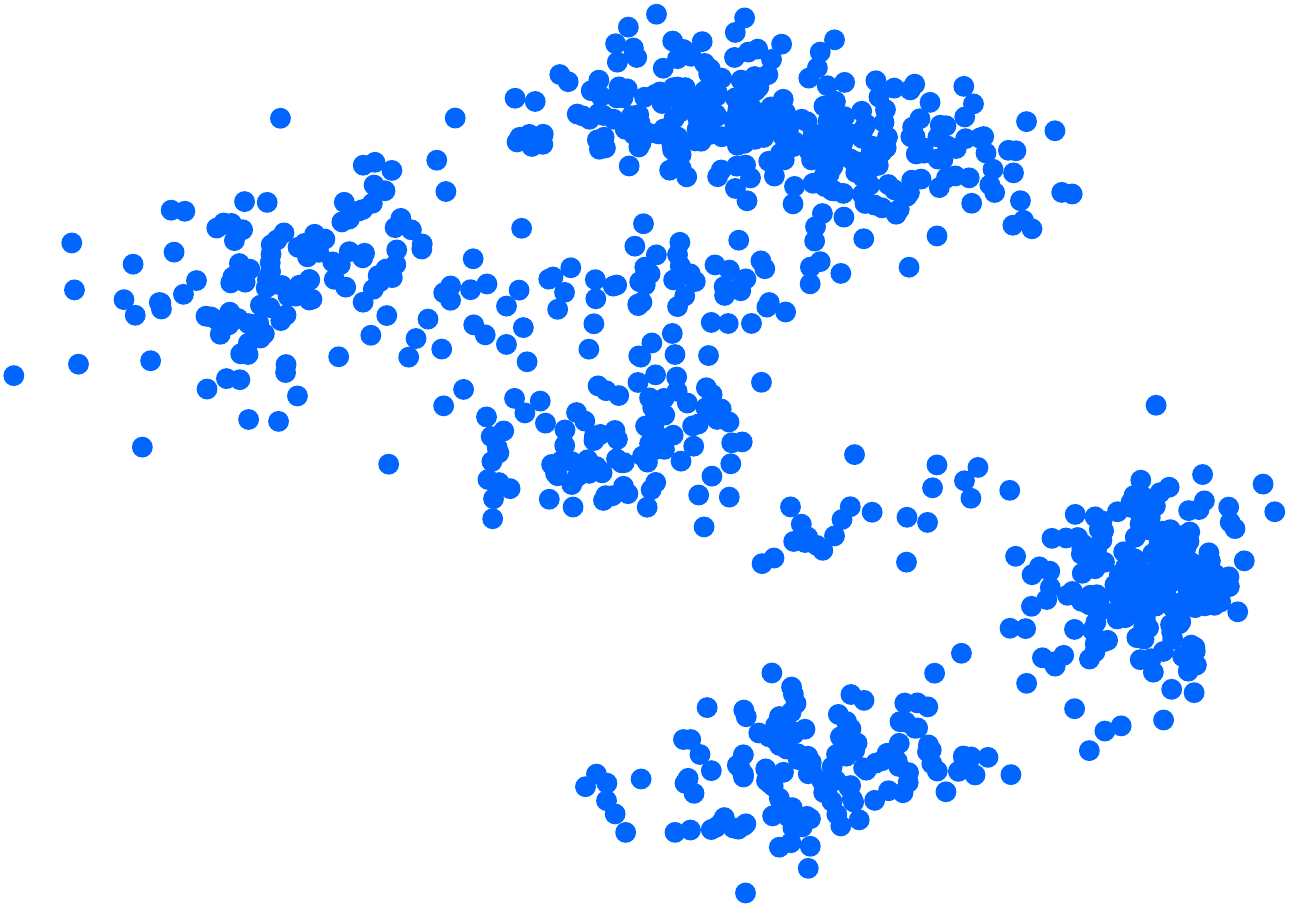}
\includegraphics[scale=0.22]{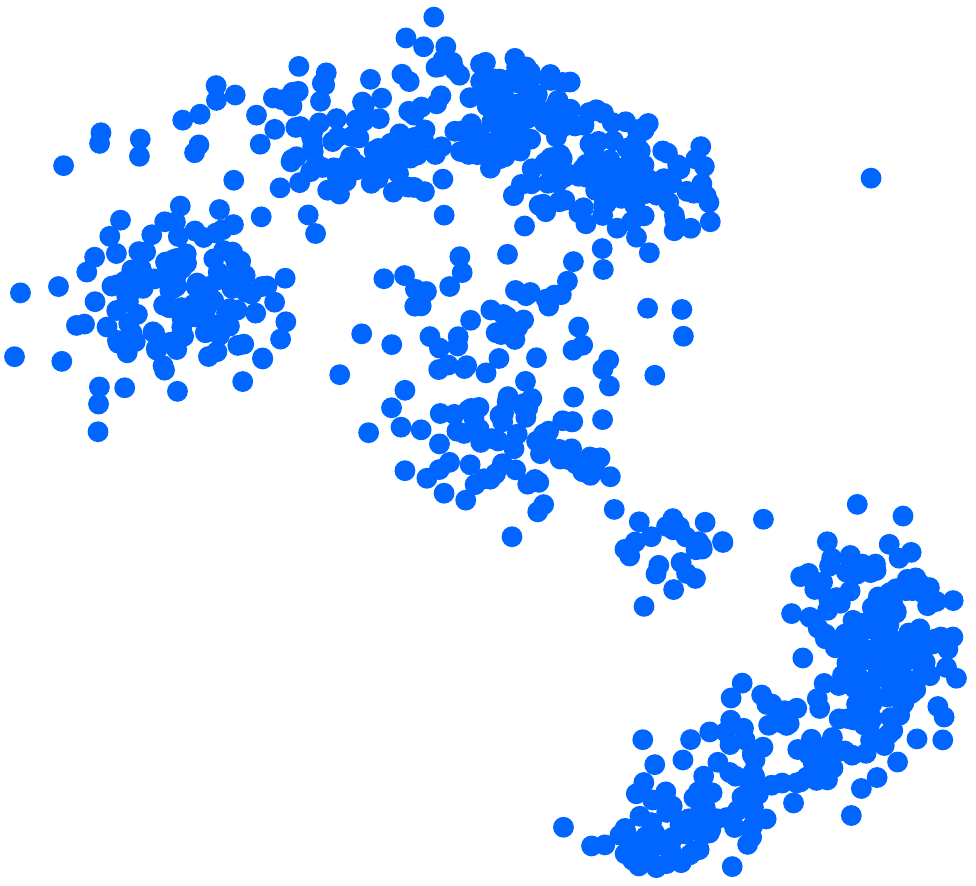}
\includegraphics[scale=0.22]{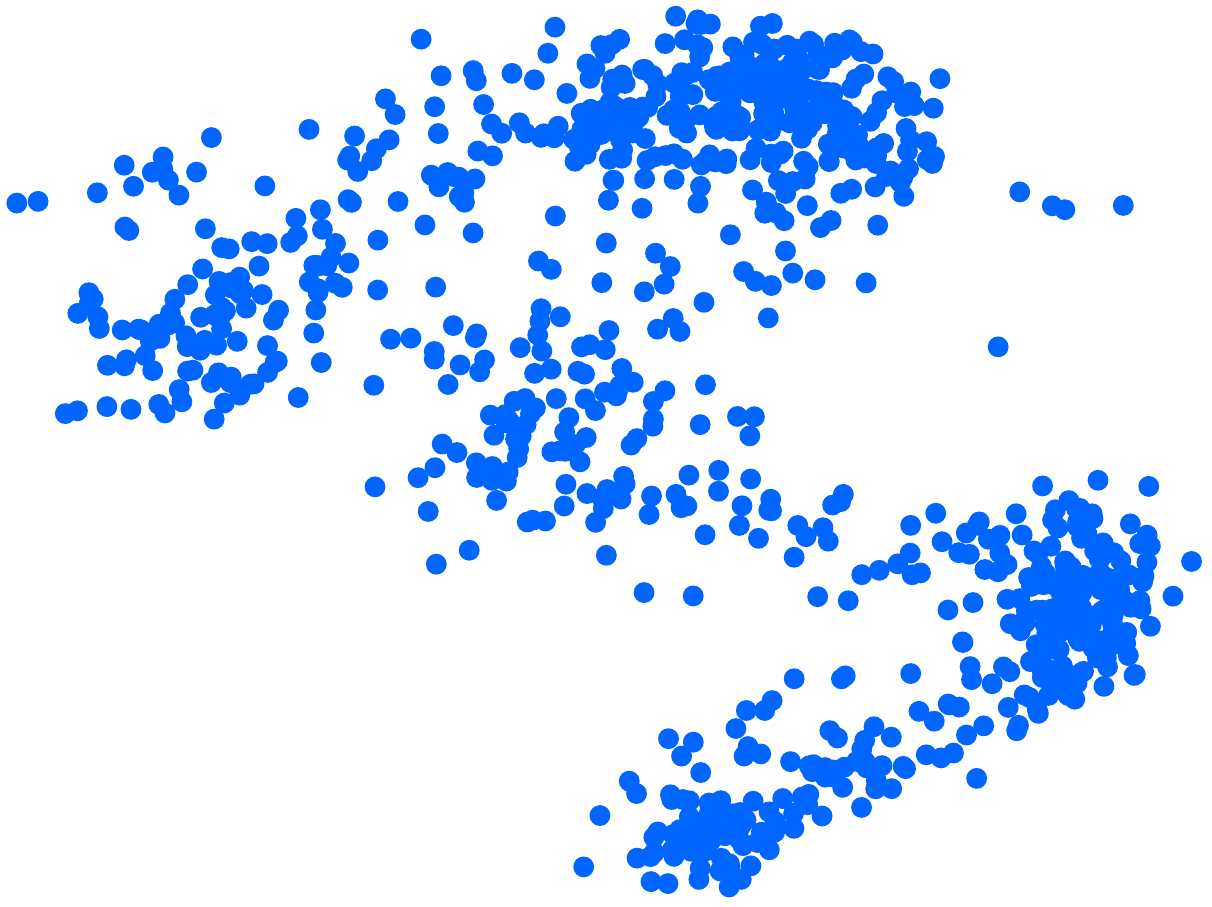}
~~~~
\includegraphics[scale=0.22]{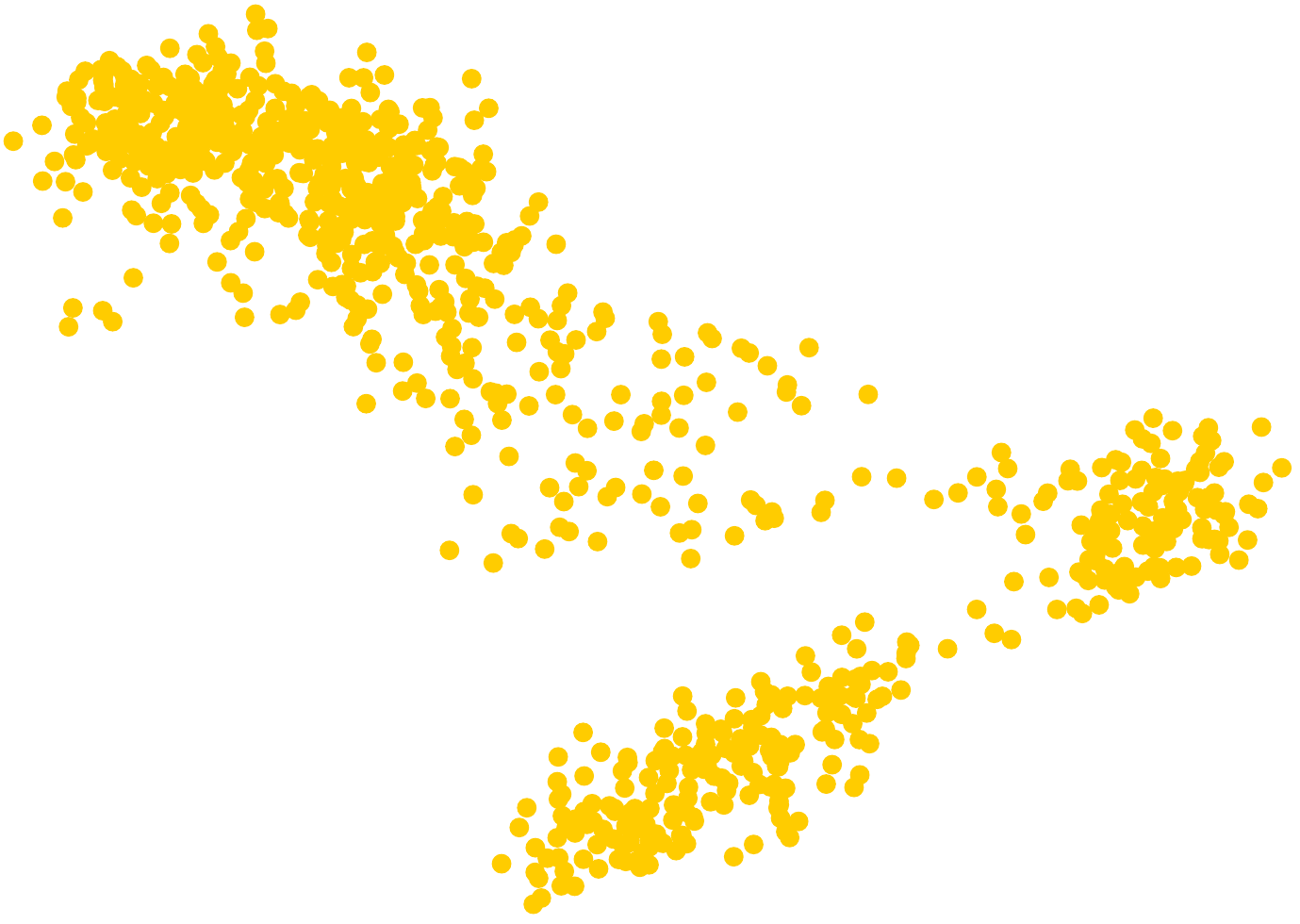}
\includegraphics[scale=0.22]{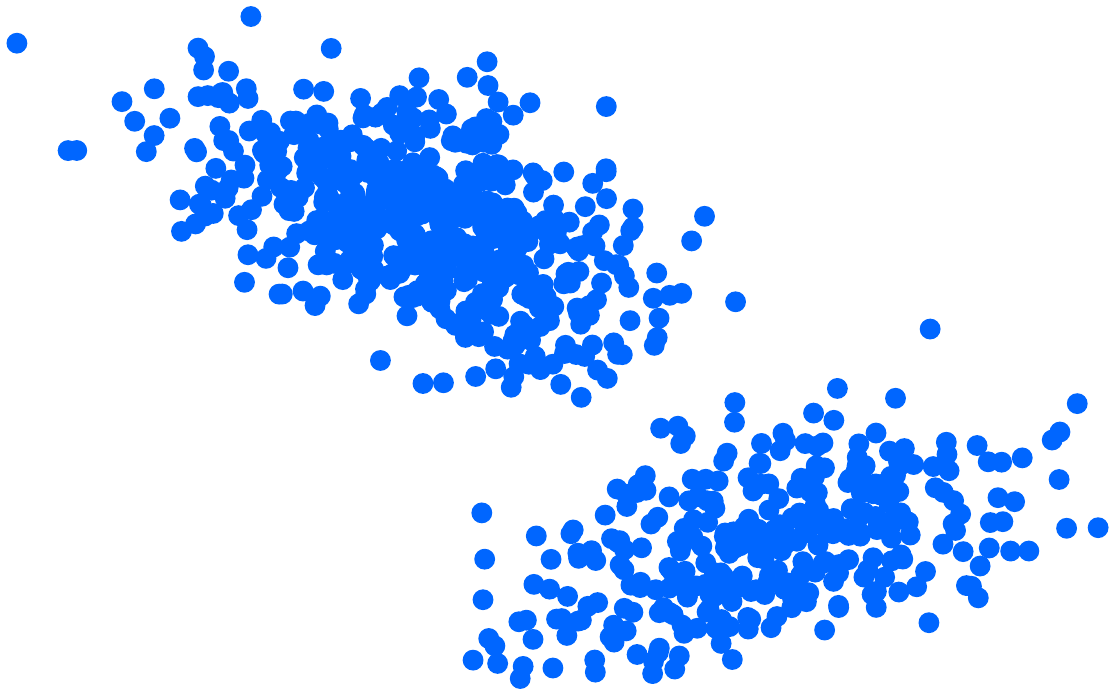}
\includegraphics[scale=0.22]{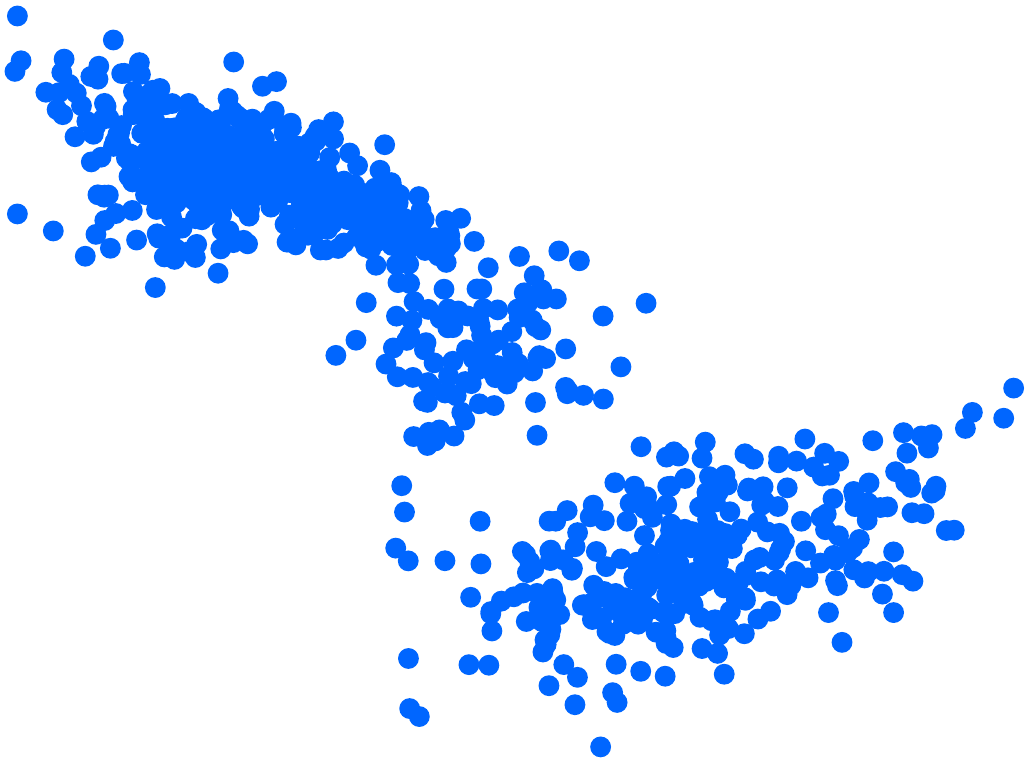}
\includegraphics[scale=0.22]{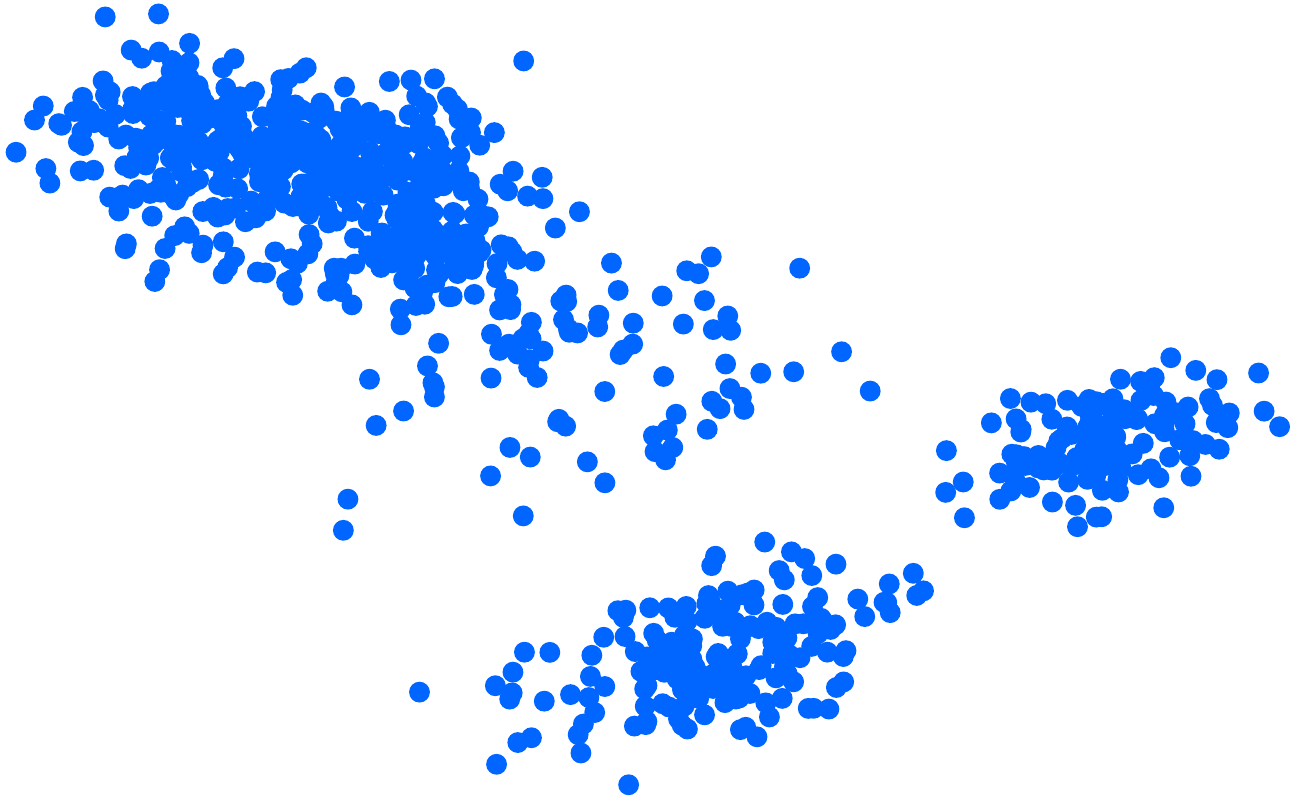}
\includegraphics[scale=0.22]{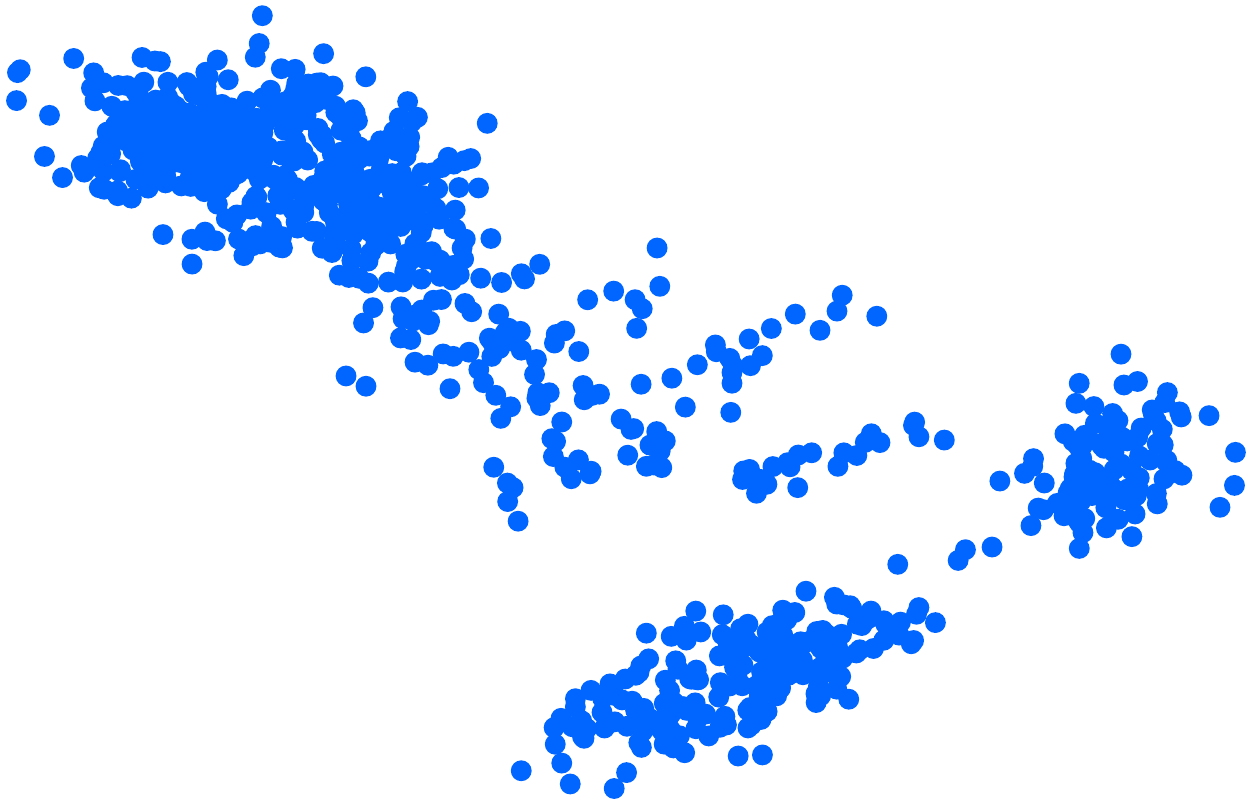}
\includegraphics[scale=0.22]{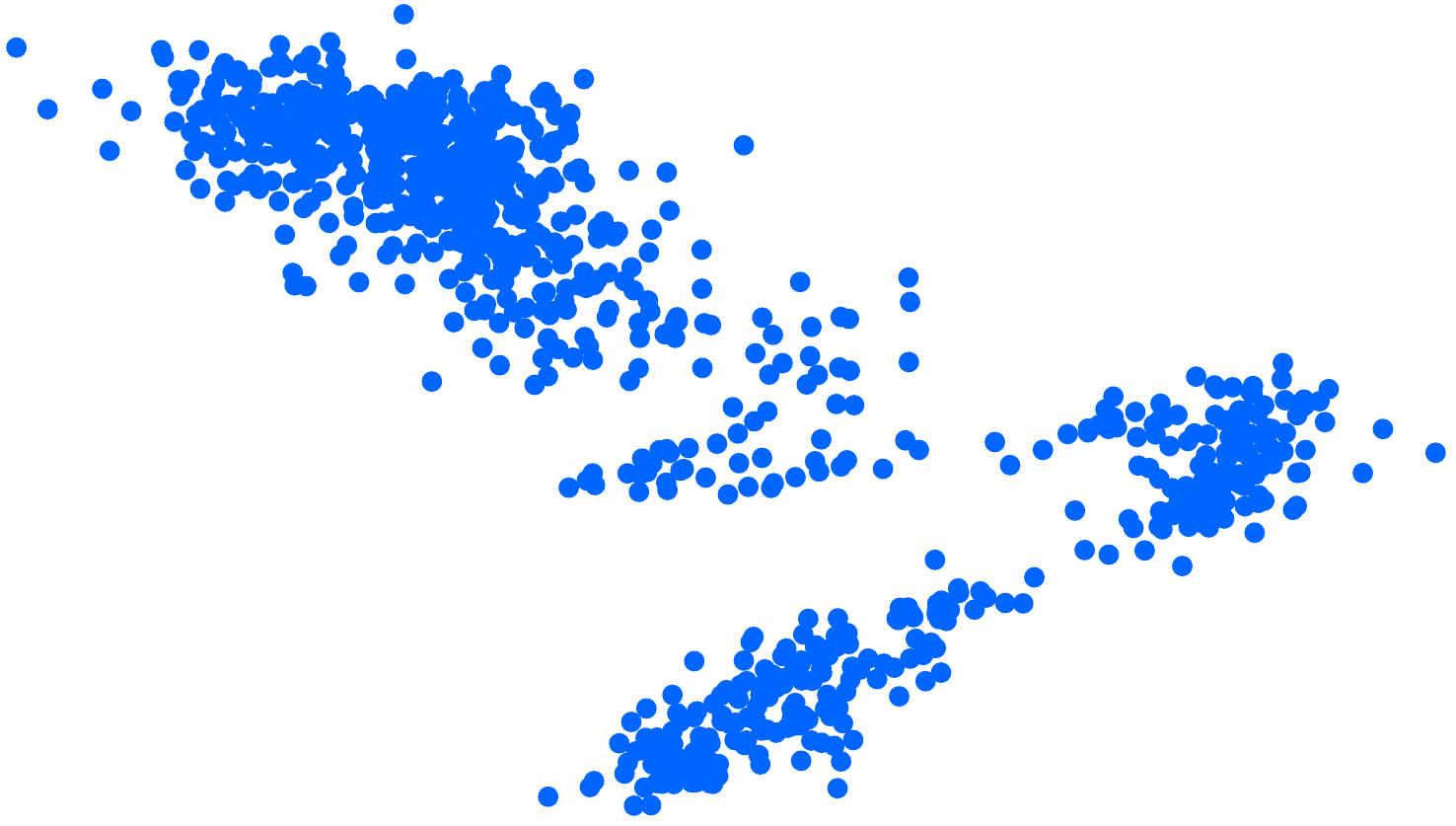}
\caption{Generated sample CC shapes for subjects without dementia (Top) and with dementia (bottom). The first column represents the original point cloud while the following columns are generated using $K=2, 4, 8, 16, 32$ GMM components.}
      \label{fig2}
\end{figure*}

\begin{figure}[!ht]
 \centering
\includegraphics[scale=0.35]{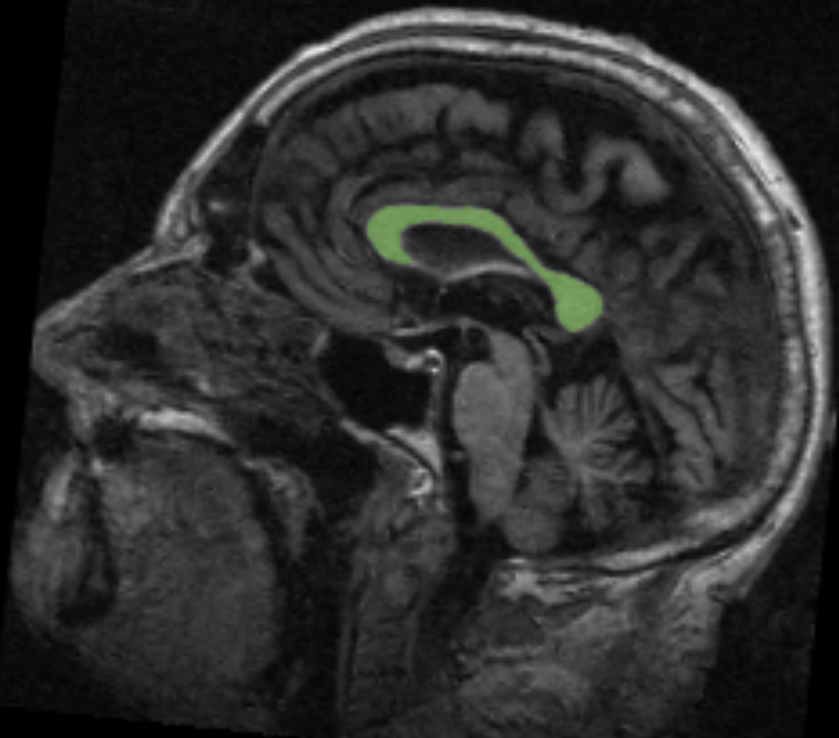}
\caption{IA sample brain MR scan overlayed with CC region. }
      \label{fig3}
\end{figure}

After rigidly registering each MR scans to the atlas, we segment out the corpus callosum region from each scan. We represent the shape of the corpus callosum as a 3D point-cloud. A sample MR scan of non-demented subject overlayed with CC highlighted in shown in Fig. \ref{fig3}.
\ \\
{\bf Interpolation experiment:} Given two CC shapes as point-clouds, we generate the interpolated shapes in between. The result is shown in Fig. \ref{fig1}. We use the two CC shapes as two endpoints of the geodesic on the product space $\mathcal{P}$. We have generated the CC point-cloud in between, i.e., with $t=0.2, 0.4, 0.6, 0.8$ respectively. Observe how the CC shapes have transformed over the geodesic. This clearly shows that our intrinsic framework can preserve the smooth transitions between CC shapes.
\ \\
{\bf Sample generation:} In this part of the experiments, we have generated CC samples from both demented and non-demented samples. Some of the generated samples for demented and non-demented classes are shown in Fig. \ref{fig2}. To see the goodness of our generated samples, we do a simple $1$-NN based classification accuracy analysis as follows.

Analogous to the training data, we have generated $33$ scans with and $36$ scans without dementia. Notice that as mentioned in Section \ref{theory}, we generate the GMM and then draw samples from it to generate the point-cloud. Given the generated GMMs, denoted by $\left\{\mathcal{N}^i\right\}$, we represent each GMM as a point on hypersphere, $\mathbf{S}^{999}$ as follows.

\begin{enumerate}
\item Draw $1000$ uniformly random samples on $\mathbf{R}^3$, denoted by  $\left\{\mathbf{y}_j\right\}_{j=1}^{1000}$.
\item For each $\mathbf{y}_j$, we compute the probability belonging to $\mathcal{N}^i$, denoted by $p_{ij}$.
\item We normalize $\left(p_{ij}\right)_{j=1}^{1000}$ to sum to $1$ and use the square root parametrization to map it on $\mathbf{S}^{999}$.
\end{enumerate}

Now that we can identify $\left\{\mathcal{N}^i\right\}$ with $\left\{\mathbf{n}_i\right\} \subset \mathbf{S}^{999}$, we use a $1$-nearest neighbor classifier to classify each generated point-cloud. Using the simple $1$-NN classifier, we can correctly classify $33$ and $33$ scans of the class demented and non-demented respectively. This results an overall $95.7\%$ classification accuracy with specificity and sensitivity to be given by $100\%$ and $91.67\%$ respectively.

\section{Conclusions}
Point-cloud helps with understanding 3D geometric shapes for medical data. But due to the lack of training samples, applicability of deep learning becomes limited in the medical image analysis. A way to overcome this limitation is by generating samples using GAN like schemes, but popular generative models are mostly suitable for natural images and hence for medical image modalities like MRI it is not appropriate to use standard GAN like schemes. In this work, we proposed a novel GMM based point-cloud generation technique and have shown that we can apply our scheme to generate new samples for  3D anatomical shapes. Experimental results have shown that we can smoothly interpolate between two given 3D shapes represented as point-clouds. Furthermore, we generated new 3D shapes and have shown that we can indeed preserve the class information in the generated samples, i.e., samples generated for demented subjects are different than that of non-demented. As a possible future direction, we like to explore the GMM based generation idea to generate other anatomical structures represented as 3D point-clouds. 
{\small
\bibliographystyle{IEEEbib}
\bibliography{references}

\begin{thebibliography}{10}

\bibitem{shin2016deep}
Hoo-Chang Shin, Holger~R Roth, Mingchen Gao, Le~Lu, Ziyue Xu, Isabella Nogues,
  Jianhua Yao, Daniel Mollura, and Ronald~M Summers,
\newblock ``Deep convolutional neural networks for computer-aided detection:
  Cnn architectures, dataset characteristics and transfer learning,''
\newblock {\em IEEE transactions on medical imaging}, vol. 35, no. 5, pp.
  1285--1298, 2016.

\bibitem{goodfellow2014generative}
Ian Goodfellow, Jean Pouget-Abadie, Mehdi Mirza, Bing Xu, David Warde-Farley,
  Sherjil Ozair, Aaron Courville, and Yoshua Bengio,
\newblock ``Generative adversarial nets,''
\newblock in {\em Advances in neural information processing systems}, 2014, pp.
  2672--2680.

\bibitem{arjovsky2017wasserstein}
Martin Arjovsky, Soumith Chintala, and L{\'e}on Bottou,
\newblock ``Wasserstein generative adversarial networks,''
\newblock in {\em International conference on machine learning}, 2017, pp.
  214--223.

\bibitem{kingma2018glow}
Durk~P Kingma and Prafulla Dhariwal,
\newblock ``Glow: Generative flow with invertible 1x1 convolutions,''
\newblock in {\em Advances in Neural Information Processing Systems}, 2018, pp.
  10215--10224.

\bibitem{qi2017pointnet}
Charles~R Qi, Hao Su, Kaichun Mo, and Leonidas~J Guibas,
\newblock ``Pointnet: Deep learning on point sets for 3d classification and
  segmentation,''
\newblock in {\em Proceedings of the IEEE Conference on Computer Vision and
  Pattern Recognition}, 2017, pp. 652--660.

\bibitem{qi2017pointnet++}
Charles~Ruizhongtai Qi, Li~Yi, Hao Su, and Leonidas~J Guibas,
\newblock ``Pointnet++: Deep hierarchical feature learning on point sets in a
  metric space,''
\newblock in {\em Advances in neural information processing systems}, 2017, pp.
  5099--5108.

\bibitem{zhou2018voxelnet}
Yin Zhou and Oncel Tuzel,
\newblock ``Voxelnet: End-to-end learning for point cloud based 3d object
  detection,''
\newblock in {\em Proceedings of the IEEE Conference on Computer Vision and
  Pattern Recognition}, 2018, pp. 4490--4499.

\bibitem{chakraborty2017statistics}
Rudrasis Chakraborty, Monami Banerjee, and Baba~C Vemuri,
\newblock ``Statistics on the space of trajectories for longitudinal data
  analysis,''
\newblock in {\em 2017 IEEE 14th International Symposium on Biomedical Imaging
  (ISBI 2017)}. IEEE, 2017, pp. 999--1002.

\bibitem{muralidharan2012sasaki}
Prasanna Muralidharan and P~Thomas Fletcher,
\newblock ``Sasaki metrics for analysis of longitudinal data on manifolds,''
\newblock in {\em 2012 IEEE Conference on Computer Vision and Pattern
  Recognition}. IEEE, 2012, pp. 1027--1034.

\bibitem{jian2010robust}
Bing Jian and Baba~C Vemuri,
\newblock ``Robust point set registration using gaussian mixture models,''
\newblock {\em IEEE transactions on pattern analysis and machine intelligence},
  vol. 33, no. 8, pp. 1633--1645, 2010.

\bibitem{fotenos2005normative}
Anthony~F Fotenos, AZ~Snyder, LE~Girton, JC~Morris, and RL~Buckner,
\newblock ``Normative estimates of cross-sectional and longitudinal brain
  volume decline in aging and ad,''
\newblock {\em Neurology}, vol. 64, no. 6, pp. 1032--1039, 2005.

\bibitem{srivastava2007riemannian}
Anuj Srivastava, Ian Jermyn, and Shantanu Joshi,
\newblock ``Riemannian analysis of probability density functions with
  applications in vision,''
\newblock in {\em 2007 IEEE Conference on Computer Vision and Pattern
  Recognition}. IEEE, 2007, pp. 1--8.

\bibitem{avants2009advanced}
Brian~B Avants, Nick Tustison, and Gang Song,
\newblock ``Advanced normalization tools (ants),''
\newblock {\em Insight j}, vol. 2, pp. 1--35, 2009.

\end{thebibliography}
\end{document}